\documentclass[10pt, twocolumn]{article}

\usepackage{times}
\usepackage{natbib}
\usepackage{fullpage}

\usepackage{tabularx}
\usepackage[table]{xcolor}

\usepackage{microtype}
\usepackage{graphicx}
\usepackage{subfigure}
\usepackage{booktabs} %
\usepackage{amssymb}
\usepackage{hyperref}
\hypersetup{colorlinks,linkcolor={blue},citecolor={blue},urlcolor={red}}
\usepackage{tikz}
\usepackage{dialogue}
\usepackage{xcolor}

\usepackage{amsmath}
\usepackage{amssymb}
\usepackage{mathtools}
\usepackage{amsthm}

\usepackage[framemethod=tikz]{mdframed}

\usepackage{listings}
\usepackage{xcolor}
\usepackage{soul}
\usepackage{pgfplots}
\usepackage{enumitem}
\usepackage[most]{tcolorbox} %

\lstnewenvironment{code}{\lstset{language=}}{}

\definecolor{completioncolor}{rgb}{1.0, 1.0, 0.4}
\sethlcolor{completioncolor}

\usepackage[capitalize,noabbrev]{cleveref}

\definecolor{wordcolor}{rgb}{0.5,0.5,1} %

\theoremstyle{plain}

\theoremstyle{definition}

\theoremstyle{remark}

\newif\ifshowcomments
\showcommentstrue
\ifshowcomments
    \newcommand{\chenhao}[1]{\textcolor{blue}{[#1 ---\textsc{CT}]}}
    \newcommand{\amit}[1]{\textcolor{brown}{[#1 ---\textsc{Amit}]}}
    \newcommand{\rn}[1]{\textcolor{orange}{[#1 ---\textsc{RN}]}}
    \newcommand{\emk}[1]{\textcolor{red}{[#1 ---\textsc{EMK}]}}
\else
    \newcommand{\chenhao}[1]{}
    \newcommand{\amit}[1]{}
    \newcommand{\rn}[1]{}
    \newcommand{\emk}[1]{}
\fi
\usepackage[textsize=tiny]{todonotes}

\definecolor{mycolor}{rgb}{0.122, 0.435, 0.698}

\newmdenv[innerlinewidth=0.5pt, roundcorner=4pt,linecolor=mycolor,innerleftmargin=6pt,
innerrightmargin=6pt,innertopmargin=6pt,innerbottommargin=6pt]{mybox}

\title{Large Language Models as Co-Pilots for\\ Causal Inference in Medical Studies}

\author{
  Ahmed Alaa \\
  {UC Berkeley and UCSF} \\
  \and
  Rachael V. Phillips \\
  {UC Berkeley} \\
  \and
  Emre K\i c\i man \\
  {Microsoft Research} \\
  \and
  Laura B. Balzer \\
  {UC Berkeley} \\
  \and
  Mark van der Laan \\
  {UC Berkeley} \\
  \and
  Maya Petersen \\
  {UC Berkeley} 
\vspace{-0.75in}
}

\date{{\small \textit{Working Paper} \today}}

\begin{document}

\maketitle

\pgfplotsset{
    colormap={wordcolor}{
        color(0)=(white)
        color(100)=(wordcolor)
    }
}

\begin{abstract}
The validity of medical studies based on real-world clinical data, such as observational studies, depends on critical assumptions necessary for drawing causal conclusions about medical interventions. Many published studies are flawed because they violate these assumptions and entail biases such as residual confounding, selection~bias,~and misalignment between treatment and measurement times. Although researchers are aware of these pitfalls, they continue to occur because anticipating and addressing them in the context of a specific study can be challenging without a large, often unwieldy, interdisciplinary team~with~extensive expertise. To address this expertise gap, we explore~the~use of large language models (LLMs) as~``co-pilot''~tools~to~assist researchers in identifying study design flaws~that~undermine the validity of causal inferences. We propose a conceptual framework for LLMs as {\it causal co-pilots} that encode domain knowledge across various fields, engaging with researchers in natural language interactions to provide contextualized assistance in study design. We provide illustrative examples of how LLMs can function as causal co-pilots, propose a structured framework for their grounding in existing causal inference frameworks, and highlight the unique challenges and opportunities in adapting LLMs for reliable use in epidemiological research.
\end{abstract}



\begin{figure}[t]
\centering
\includegraphics[width=3in]{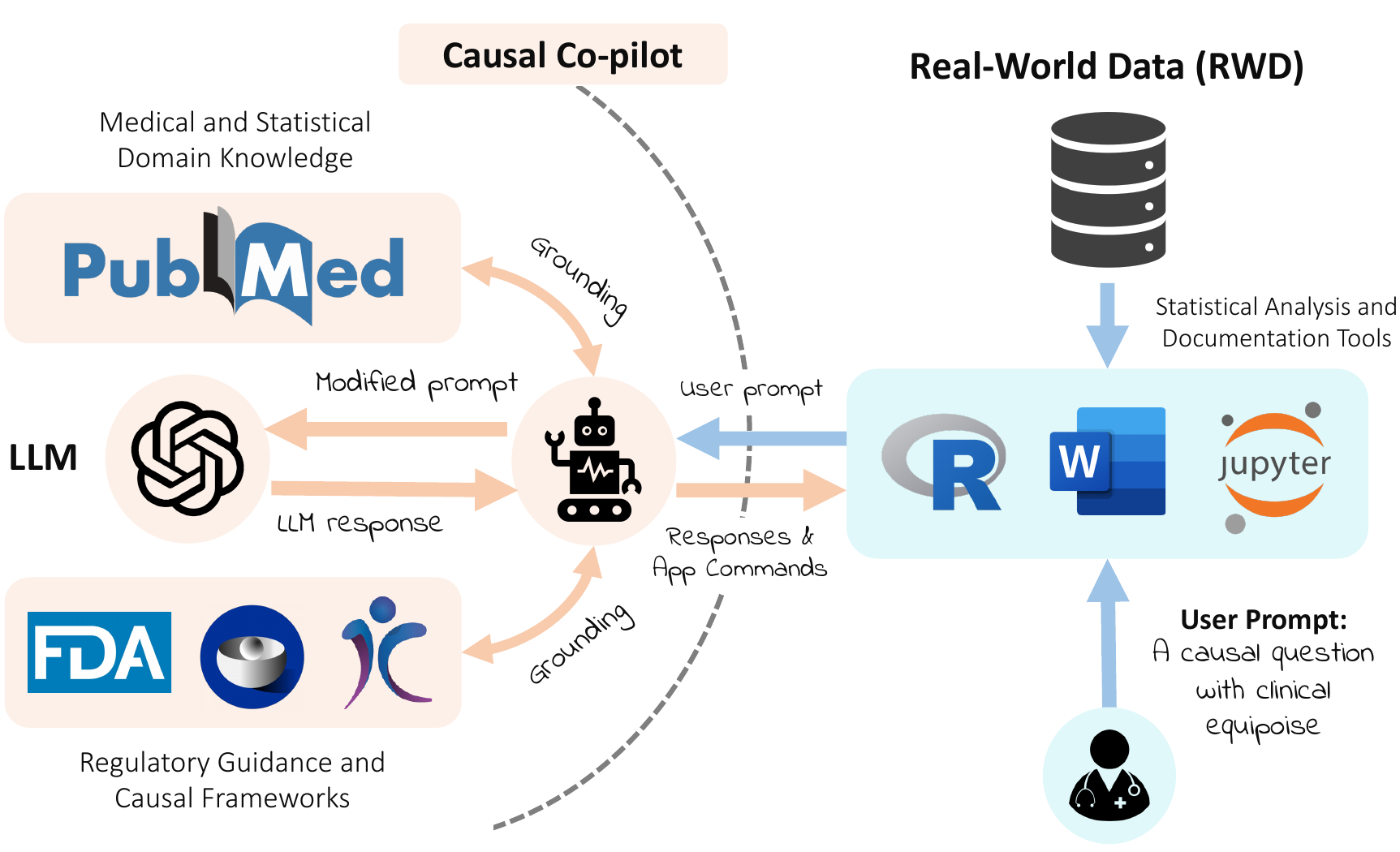}
\vspace{-.15in} 
\caption{{\footnotesize {\bf Overview of the Medical \textit{Causal Co-Pilot} Framework:} The causal co-pilot engages with input prompts from users (e.g., clinical and biopharmaceutical researchers) related to a causal question with clinical equipoise, such as determining if a drug causes an increase in the incidence of a future adverse event. Users provide contextual information on the RWD being utilized. The causal co-pilot~then~refines prompt specificity by {\it grounding} it in clinical and statistical domain knowledge, as well as regulatory guidance and analytic frameworks for causal inference. This collaborative interaction results in a rigorous and transparent study aimed at addressing the causal question using the RWD at hand.}}
\label{Fig1}
\vspace{-0.25in}
\end{figure}

\section{Introduction}
\label{Sec1}
Real-world data (RWD) play a vital role in advancing medical knowledge, from assessing correlates and causes of human disease to evaluating drug effectiveness and safety in routine clinical practice settings (\cite{black1996we, yuan2023comparison}). RWD can come from electronic health records (EHRs); insurance claims and billing activities; disease and medical product registries; and other data sources that are informative of patient health status, such as digital health technologies. Studies that are designed to integrate RWD (such as observational studies (OS), pragmatic trials, and externally controlled trials) are often used when randomized controlled trials (RCTs) present ethical concerns or feasibility challenges. This can occur (i) when studying populations that are vulnerable or unable to give informed consent (such as children, cognitively impaired individuals, or severely ill patients); (ii) in studies of diseases with high mortality rates and no existing effective treatments, as withholding potential treatments for the sake of randomization could be seen as unethical; or (iii) in public health emergencies where delaying treatment for the sake of a randomized trial could result in significant harm or death. Studies also use RWD when it is of interest to better understand longer term outcomes or patient populations that are more representative of routine care (\cite{concato2012randomize}). 

In contrast to RCTs, studies with RWD have inherent limitations and complexities (such as confounding, selection bias and misalignment between measurements) that may introduce biases if they are not rigorously addressed (\cite{stroup2000meta, hannan2008randomized, shrank2011healthy, fox2024s}). Notably, without~randomization~of~treatment, {\it confounding} persists; these studies are susceptible to identifying associations that do not necessarily imply causation (\cite{gerstein2019real}).~Historical~instances~have demonstrated that inadequate study design, analysis, or misinterpretation of results have led to publications that could mislead public heath directives (\cite{kral2005flaws}). Such studies are even occasionally cited in the media, potentially fostering misconceptions among the public, particularly during health crises such as the COVID pandemic (\cite{lyons2011spread, alexander2020covid, jung2021methodological}). 

On the other hand, when studies using RWD are carefully designed and rigorously analyzed they offer the potential to advance medicine and improve clinical care. In particular, RWD studies that meet evidentiary standards can contribute to drug development and be used to support regulatory decision-making (\cite{concato2020randomized}). For example, in 2021, the Center for Drug Evaluation and Research (CDER) of the US Food and Drug Administration (FDA) granted approval to a medication following a study using RWD. This approval pertained to the use of tacrolimus (Prograf) along with other immunosuppressants to prevent organ rejection in lung transplants. The decision was based on an analysis comparing data from a well-established registry with data on historical controls. This marked the first time CDER acknowledged an OS as a sufficient, well-controlled investigation that provided primary support for a finding of substantial treatment effectiveness evidence (\cite{concato2022real}). Legislation enabling the use of RWD in regulatory decision-making has now been passed across the globe.

Despite increased recognition of the potential of RWD and the substantial efforts to use it to advance medicine, mechanisms for following best practices and controlling the quality of RWD studies have not kept pace. Indeed, it is an arduous task to design studies that adhere to these standards. It requires (i) knowledge of regulatory guidance documents; (ii) detailed and accurate descriptions of the RWD source and its reputability; (iii) objective comparisons of different methodologies for generating rigorous causal inferences in the absence of randomization.; (iv) respecting the RWD complexity via sophisticated and robust statistical and causal modeling; and (v) thorough discussion of the implications of the results, limitations, and any assumptions and their plausibility. To accomplish this, subject matter expertise across several domains is necessary, including medicine, public health, epidemiology, statistics, and causal inference. In practice this typically involves multiple iterations across teams of interdisciplinary experts. Key judgment calls may not be fully vetted or transparently reported, and the burden is placed entirely on the human investigator team with little to no technical support. 

In this paper, we outline a new research direction exploring how large language models (LLMs) can collaborate with researchers to improve the quality of medical studies based on RWD. Our key thesis is that, while pitfalls in RWD studies are diverse and numerous, many are avoidable and can be addressed during study design and analysis. Because LLMs encode domain knowledge across various fields (\cite{singhal2022large}), they can emulate multidisciplinary expertise to help detect these pitfalls within the specific context of a study and engage with researchers through natural language interactions to refine study design and avoid these issues. Throughout this~paper,~we~provide illustrative examples of how LLMs can function as causal co-pilots, propose a structured framework for integrating LLMs with existing causal inference methodologies, and highlight the unique challenges and opportunities in adapting LLMs for reliable use in epidemiological research.

In the next Section, we provide a brief overview of the different types of RWD studies and existing frameworks for RWD-based causal inference. Readers versed in this domain may opt to proceed directly to Section 3. 

\vspace{-0.1in}
\section{Preliminaries}
\label{sec-landscape}
\subsection{Landscape of Medical Studies}
\label{sec-landscape-studies}
Contemporary discourse often simplifies studies into ``randomized'' or ``observational'' types, neglecting the wide variety of data and designs that exist. Regulatory agencies across the globe have worked to rectify these misunderstandings and provide a more nuanced categorization of studies involving RWD (\cite{ich2023framework}). In Figure~\ref{fig:landscape}A, we illustrate the spectrum of studies that use RWD and generate real-world evidence (RWE, clinical evidence about the usage and potential benefits or risks of a medical product derived from analysis of RWD (\cite{fda2018framework})). Figure~\ref{fig:landscape}A depicts the spectrum of medical studies along with the degree of RWD reliance across different types of studies. Figure~\ref{fig:landscape}B highlights the key challenges involved in designing RWD studies. We note that these challenges are not necessarily more prevalent with increasing reliance on RWD. For instance, a well-designed observational study may present fewer challenges than an externally controlled trial with notable differences between the external control and single-arm RCT subjects.

\vspace{-0.1in}
\subsection{Frameworks for RWE Generation} 
\label{sec-landscape-causal}
\begin{figure}[h]
\centering
\includegraphics[width=3.15in]{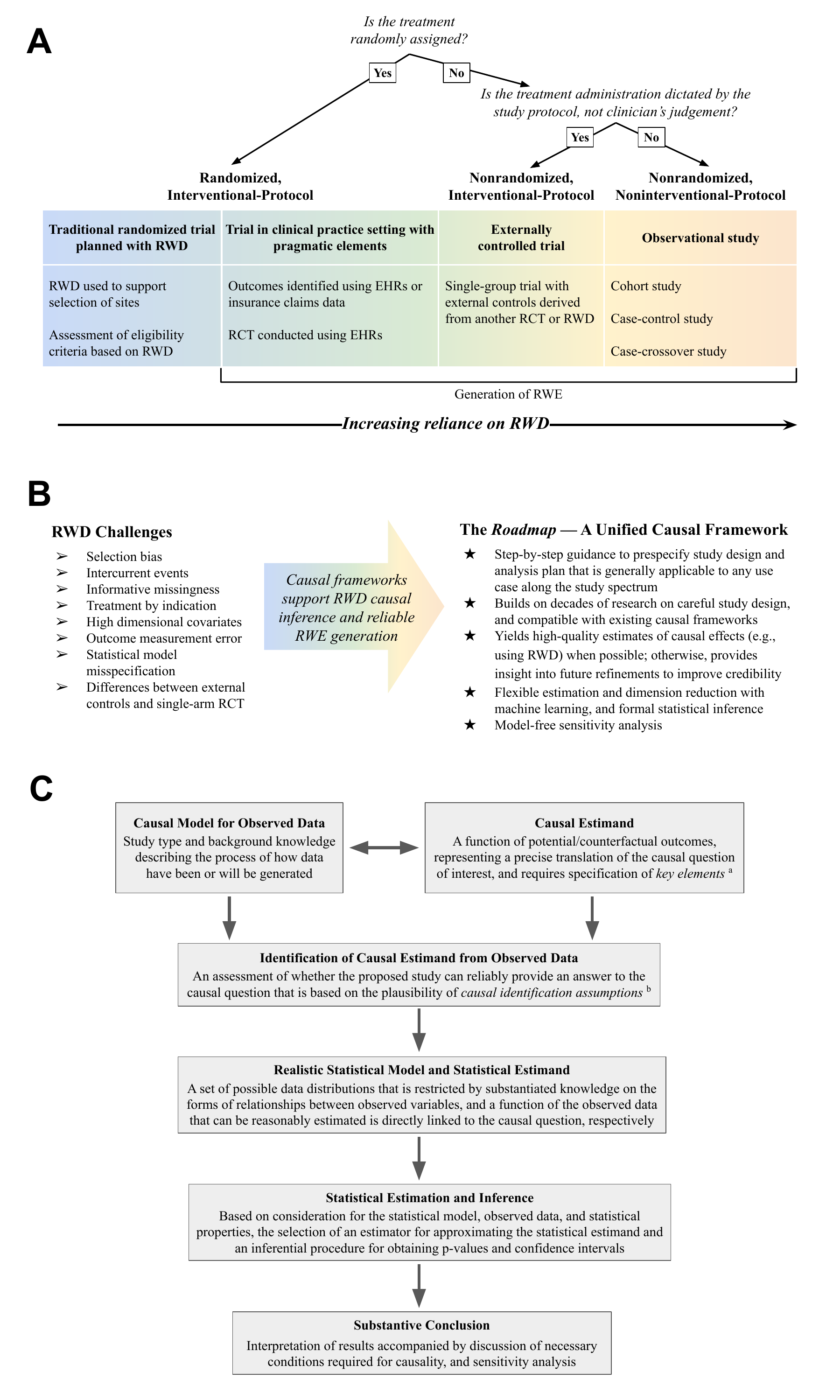}
\vspace{-.3in} 
\caption{{\footnotesize Landscape of medical studies and RWE frameworks. Panel A is adapted from \cite{concato2020randomized} and panel C is adapted from \cite{dang2023causal, petersen2014causal}.}} 
\label{fig:landscape}
\vspace{-0.2in}
\end{figure}

Designing RWD studies involves numreous complexities (Figure~\ref{fig:landscape}B) that make generating valid RWE challenging. These complexities also complicate the clear communication of findings in scientific publications, to the extent that many medical journals restrict the use of notions of causality to RCTs (\cite{dahabreh2024causal}). To address these issues, researchers have proposed structured approaches to designing and communicating RWD studies to ensure the validity and clarity of their causal conclusions. These frameworks provide guidance on defining and reporting the causal question being studied, the causal quantity being estimated, the study design, the assumptions made, and the use of RWD to answer the causal question (\cite{dang2023causal, petersen2014causal}).

The {\it Causal Roadmap} is an example for such a structured framework that we have previously proposed in \cite{dang2023causal}. The roadmap is an explicit, iterative step-by-step guide for investigators to follow in the pre-specification of study design, causal inference and statistical analysis plans. It addresses a wide range of questions and handles various RWD study types within a single framework. Using the Causal Roadmap helps in (i) detailing key elements of a study design and analysis plan likely to yield high-quality evidence; and (ii) reaching an objective conclusion that producing the required level of medical evidence is presently unfeasible, while also identifying what data is necessary for future credible evidence generation. 

\vspace{-0.1in}
\section{How can LLMs Help Avoid Flaws in RWD Studies?} 
\label{Sec2} 

\begin{mybox}
{\color{black}{

{\bf TL;DR:} LLMs possess the capability to analyze RWD study designs and results, both in textual and visual formats, while also discerning potential biases within these studies. However, off-the-shelf LLMs struggle to provide accurate and pertinent responses without explicit user instructions and explanations of general principles that guide their understanding of bias sources.

}}
\end{mybox} 

Despite the availability of structured guidance for designing RWE studies, researchers often struggle to apply this abstract guidance to specific studies. This difficulty arises because each step in the study design process can present multiple potential flaws, and reasoning through these issues often requires context-specific and domain-specific knowledge. For instance, the initial step of a RWD study typically {\it ``defines a causal question \& estimand''}. As we will discuss in this section, there are many ways researchers can define an estimand that does not address the causal question of interest, or they might select a causal question that is not clinically relevant. In this section, we demonstrate how LLMs can, in principle, serve as ``causal co-pilot'' tools for contextualizing the application of structured frameworks for causal inference to the RWD study at hand. However, we also show that the zero-shot application of widely-used LLMs does not fulfill this~role~effectively.

We begin by motivating the causal co-pilot role of LLMs through a demonstration of how an LLM can collaborate with researchers to intercept problems in designing studies using RWD. We showcase how an LLM co-pilot can help identify and avert flaws, drawing on those from three well-known historical instances (Sections \ref{Sec21}, \ref{Sec22}, and \ref{Sec23}) where findings of observational studies (OS) were later refuted. In this section, we focus on OS due to their historical prevalence in clinical research. We discuss the role of LLMs in other types of RWD studies in the next section.

\vspace{-.1in}
\subsection{Clarifying the Causal Question}
\label{Sec21}
Generating valid evidence from RWD requires specifying a precise causal question---a query regarding the cause-effect relationship between a possibly time-varying exposure or treatment on an outcome for a particular target population of interest. In practice, this process can involve multi-disciplinary teams collaborating for months to pinpoint an estimand that (i) is clinically meaningful, and (ii) can be effectively answered using available data. 

An LLM co-pilot can assist researchers in defining the causal question for the study at hand. The user-LLM interaction on the causal question can encompass various aspects of the causal estimand, including defining eligibility criteria, follow-up timing, outcome variables and treatment protocols, and definition of baseline study time. Responding to these questions, the co-pilot can leverage its encoded medical domain knowledge (\cite{singhal2023large}) to contextualize its responses in a way that mirrors a multi-disciplinary collaboration. Below, we describe a notable instance of misleading findings from OS that were later attributed to poor formulation of the causal query, and then we discuss how an LLM co-pilot could have provided guidance to intercept the errors underlying the study design.

\textit{\textbf{Does postmenopausal hormone therapy have cardiovascular benefits?}}
~In~the~1980s~and 1990s, several OS examined the impact of hormone replacement therapy (HRT) on future coronary heart disease (CHD) events in postmenopausal women, with results suggesting a cardiovascular benefit from HRT (\cite{barrett1998hormone}). These findings were later contradicted by multiple RCTs, including the Women's Health Initiative study (\cite{manson2003estrogen}), which even stopped early because of an {\it increased} risk of CHD and breast cancer in women treated with HRT (\cite{boardman2015hormone}). Medical guidelines and recommendations were updated in light of these RCT findings (\cite{nih-hrt-chd}).

An OS postmortem revealed that the biases exhibited where the result of~a~flawed~conceptualization of the underlying ``causal question'' (\cite{hernan2008observational}). Firstly, the OS excluded women who initiated HRT and then suffered nonfatal myocardial infarction within the first 2 years, leading to selection bias that underestimates short-term risk of HRT on CHD event incidence. Additionally, the OS compared incidence of CHD events in women who were currently on HRT and those who were never on HRT (prevalent users), whereas the RCT compared HRT initiators and non-initiators (incident users). Comparing current users may be less clinically significant and is more prone to bias here, as women already on HRT for a long time are less likely to be susceptible to risk.

\begin{figure*}[t]
\centering
\includegraphics[width=6.25in]{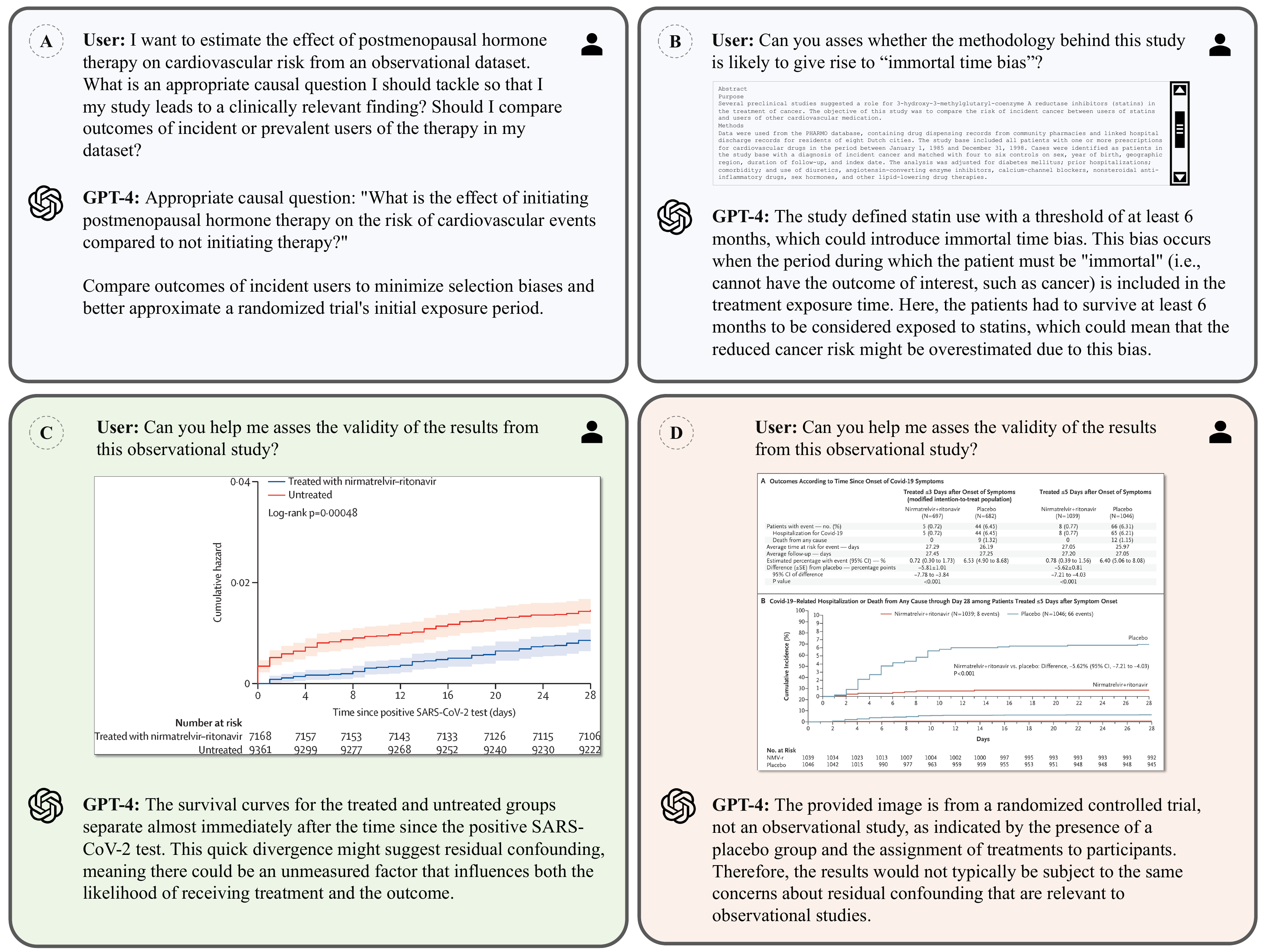}
\vspace{-.1in} 
\caption{{\footnotesize {\bf Demonstrating the Capabilities of LLMs as Causal Co-pilots:} We assess the ability of GPT-4 to analyze the design and results of the observational studies in Sections \ref{Sec21}, \ref{Sec22}, and \ref{Sec23} through various forms and modalities of interaction with a human user. {\bf Panel A:} The user directly queries the LLM co-pilot about the appropriate causal question for designing a study on the impact of postmenopausal hormone therapy on cardiovascular risk. {\bf Panel B:} The user supplies the entire text of the published study in \citet{graaf2004risk} and asks the LLM co-pilot to evaluate if the design is prone to immortal time bias. {\bf Panels C \& D:} The user~requests~the LLM to assess the validity of study results based on visual inputs in the form of cumulative hazard curves. (Visual inputs in Panels C \& D are derived from Figure 1 \& 2 in \citet{aggarwal2023real} and \citet{hammond2022oral}, respectively. Copyright clearances and permissions to reuse both figures were obtained from the original publishers.)}} 
\label{Fig2}
\vspace{-.05in}
\rule{\linewidth}{0.35pt}
\vspace{-0.3in}
\end{figure*}

\textbf{An LLM-based causal co-pilot can account for various aspects of the causal estimand and provides sound reasoning for its specification.} The lesson learned from the OS on the effects of HRT is that an imprecise definition of the causal question can yield misleading or even clinically irrelevant findings. In Figure \ref{Fig2} (Panel A), we query an LLM model (GPT-4) about the appropriate causal question for estimating the effect of HRT on incidence of CHD events. We find that, with a careful choice of prompts, the model shows sound reasoning on how to specify the causal question in the context of the study. {\bf On~the~flip~side}, without user prompts providing cues on bias sources, an off-the-shelf LLM typically offers generic responses and fails to preemptively alert the researcher regarding relevant bias considerations for the study at hand.

\vspace{-.1in}
\subsection{Critiquing Study Design}
\label{Sec22}
In contrast to RCTs where interventions are randomized and follow strict protocols, OS often require retrospective decisions on defining treatments, follow-up~time, and outcomes. Even with a sound causal question, poor design and analysis choices in OS can introduce problems in the definition of the estimand and biases in the corresponding estimates, ultimately producing potentially unreliable results that must be interpreted with caution. 

The causal co-pilot can collaborate with researchers to enhance study design by critiquing \textbf{initial} design choices made by researchers and identifying potential sources of bias, thereby intercepting a chain of problems that may result from issues with the definition of the estimand itself. Below, we describe a historical instance of this chain effect; specifically, a flawed OS design which induced immortal time bias. We then discuss how the LLM causal co-pilot could have intercepted this issue at the study design phase.

\textit{\textbf{Does statin use reduce risk of future cancer?}}
Statins are medications that reduce LDL-cholesterol. ~In~the early 2000s, several OS suggested a substantial decrease in the risk of various cancers, including colorectal and lung cancer, associated with statin use (\cite{graaf2004risk, poynter2005statins, khurana2007statins}). These results were later contradicted by RCTs that found no impact of statins on cancer risk (\cite{cholesterol2012lack, dale2006statins}).

A~critical review of the OS identified~that~the~bias originated primarily from a failure to establish a suitable ``time zero'' or ``baseline'' (starting point from which all measurements are taken). That is, there was a lack of synchronization between patient eligibility criteria and treatment assignment times (\cite{dickerman2019avoidable}). This misalignment induces a form of selection bias known as ``immortal time bias'' (a period during which the outcome~cannot~occur is incorrectly included in the analysis), which can lead to inflated effect estimates (\cite{yadav2021immortal}). 

\textbf{An LLM causal co-pilot can intercept bias and provide plausible explanation regarding its source.}~In~Figure~\ref{Fig2} (Panel B), we provide the LLM with the entire text of the OS in (\citet{graaf2004risk}) and ask it to check~for~any~aspects of the study design that may lead to immortal time bias. The LLM highlights the definition of statin use in the study as a potential issue, which is a plausible explanation of the source of bias based on the study details. {\bf However}, similar to the previous example, the ability of the LLM to detect bias in this study relies on receiving cues on what to examine; otherwise, its responses are typically generic.

\vspace{-.1in}
\subsection{Interpreting Study Results}
\label{Sec23} 
Another way the causal co-pilot can collaborate with researchers is by having the LLM critically analyze study results to uncover signs of residual confounding or other biases. Below, we illustrate how this can be done, drawing on various OS that examined the effectiveness of Nirmatrelvir/Ritonavir (brand name ``Paxlovid'') in the more recent omicron variant surges of the COVID-19 pandemic.

\textit{\textbf{Does Paxlovid reduce the risk of hospitalization and mortality in COVID-19 patients?}}
During the COVID-19 pandemic, the timeliness and urgency of policy decisions necessitated the use of RWD to gather evidence on pharmaceutical interventions. In 2021, Paxlovid received emergency use authorization from the FDA for COVID-19 treatment.~The~primary~data supporting the authorization was based on the EPIC-HR trial---an RCT that studied the efficacy of Paxlovid in non-hospitalized adults with a confirmed SARS-CoV-2 infection (\cite{hammond2022oral}). Following the EPIC-HR trial, various OS subsequently examined the effectiveness of Paxlovid in the more~recent~omicron variant surges. One OS~found~that~the treatment was associated with improved 28-day outcomes compared to no antiviral treatment (\cite{aggarwal2023real}). 

A~common indicator of residual confounding~in~OS of COVID-19 vaccines and drugs is the premature separation of survival curves for treated and control patients, occurring before the mechanism of action for the drug could influence outcomes (\cite{mohyuddin2023detecting}). In this study, the reported cumulative hazard curves separated at time zero, which is implausible given that the effect of Paxlovid takes place a few days after administration.

\textbf{An LLM co-pilot can scrutinizes study results and uncovers biases based on visual inputs of scientific graphs.} In Figure \ref{Fig2} (Panels C \& D), we show the LLM images containing the results of the OS of Paxlovid in \citet{aggarwal2023real} (Panel C) and the results of the EPIC-HR trial, with the LLM blinded to the fact that it was an RCT (Panel D). The LLM is explicitly instructed to examine any raw data provided by the user and check for indications of residual confounding. The LLM responses to the images demonstrated an ability to process visual input, comprehend the displayed data, identify the intervention being studied, and draw conclusions based on its knowledge of the mechanism of action of Paxlovid. In this example, the LLM suspected a residual confounding bias in the results of the OS. {\color{black}{Similar to the previous examples, we included instruction for the LLM on how to look for residual confounding in the study results  (see Appendix).}} Notably, in Panel D the LLM recognized that the data comes from an RCT rather than an observational study as the user ``mistakenly'' indicated in the prompt. It did so by recognizing the existence of a placebo group in the input figure.

\vspace{-.1in}
\section{The Causal Co-Pilot Framework} 
\label{Sec3} 
The examples in Section \ref{Sec2} illustrate how LLMs can function as causal co-pilots by actively collaborating with researchers to refine study design, critique existing studies, or identify potential biases in study results. Importantly, the applications of these LLM co-pilots go beyond OS to various types of studies using RWD. In this section, we explore the potential role of LLMs as causal co-pilots in the current landscape of these studies (Section \ref{sec-role})~and~present their envisioned system architecture (Section \ref{sec-architecture}).

\vspace{-.1in} 
\subsection{Role of LLM Co-Pilots in RWD Studies}\label{sec-role} 
The potential of LLMs to serve as co-pilots in medical studies hinges on two critical factors: (i) pretraining on diverse internet data, which enables LLMs to encode extensive medical and statistical domain knowledge (\cite{jin2019pubmedqa, singhal2023large}), and (ii) the capability to reason and communicate with users in natural language, which enables LLMs to contextualize its encoded domain knowledge (\cite{wei2022emergent, kojima2022large, pal2022medmcqa, nori2023capabilities}). As depicted in the spectrum of studies in Figure \ref{fig:landscape}, the increasing reliance on RWD introduces more design choices for researchers, which raises the likelihood of poor decisions and calls for the co-pilot's involvement in the study design process.

In the previous section, we explored how an LLM co-pilot can contribute to the design and interpretation of OS, which is positioned at the far end of the study spectrum. The co-pilot's applicability is more broad, however; studies that only partially rely on RWD can benefit from this framework. For example, in externally controlled trials, data for the control arm may be collected from RWD sources such as disease registries, electronic health records (EHRs), and medical claims (\cite{jahanshahi2021use}). This introduces additional {\it design considerations}, such as RWD population characteristics, treatment attributes, determination of ``time zero'', and outcome assessment. Additionally, it raises {\it data considerations}, such as assessing the comparability of trial arms. The LLM co-pilot can assist researchers in selecting RWD for the control arm in an externally-controlled trial by using knowledge on the timing, frequency of data collection, and care patterns across different RWD sources (e.g., EHR vs. disease registries vs. medical claims). 

\vspace{-.1in} 
\subsection{Causal Co-Pilot System Architecture}\label{sec-architecture}
To~fulfill~the vision described above, we envision the architecture of the causal co-pilot resembling that~of~other~assistant tools such as Github Copilot (\cite{nguyen2022empirical}) (Figure \ref{Fig1}). The causal co-pilot should~adeptly~process~user input prompts from diverse stakeholders, such as clinical and biopharmaceutical researchers, addressing causal questions with clinical equipoise. For instance,~the~user~may inquire about the scenarios discussed in Section \ref{Sec2}. Additionally, users can specify details about the RWD in use, such as its origin from an EHR or a claims database. 

Such key attributes of the study design (presented in Section \ref{sec-landscape}) provide key inputs to the co-pilot as the relevant guidance, ethical concerns, and implications will vary depending on this context. The co-pilot will then refine the specificity of the user prompt by grounding it in clinical, causal and statistical domain knowledge, as well as potential in regulatory guidance for RWE studies, as we discuss in more detail in the next Section. This grounding ensures that the underlying LLM consistently delivers relevant and accurate responses that are useful for the researcher and tailored to the specific medical study. The co-pilot tool can integrate within commonly used tools for conducting and documenting statistical analyses, such as RStudio and Jupyter notebooks. In addition to assisting in study design steps, it can automatically generate documentation of study design parameters and rationale behind design choices. This could enhances the transparency of these studies and mitigate concerns regarding data dredging (\cite{smith2002data}).

In this framework, one can imagine the LLM co-pilot as a ``research agent'' that undertakes different ``actions'' throughout the research process, similar to \cite{boiko2023autonomous}. In this sense, the action space of the co-pilot would encompass typical atomic tasks within the research process, such as \textit{``Search the web"}, \textit{``Write code"}, \textit{``Document"}, and \textit{``Check FDA Guidance"}. The co-pilot can operate with a fixed system prompt, following a predetermined logic in selecting actions during user interactions. 

\vspace{-.1in} 
\section{Developing a Causal Co-Pilot:\\ Research Directions \& Challenges}
\label{Sec4} 

\begin{mybox}
{\color{black}{

{\bf TL;DR:} To make LLMs effective as causal co-pilots, we must go beyond standard training methods.~The~challenge lies in the presence of flaws in many published studies that would be used for training, as well as the lack of real-world instruction-tuning data. Developing a causal co-pilot requires novel techniques grounded in causal inference frameworks and regulatory guidelines.

}}
\end{mybox} 

In the previous sections, we provided an overview of the medical studies landscape and outlined the potential use cases for LLM co-pilots in supporting the design of these studies. LLM-based assistance tools have found applications in various domains, including medical conversational models for diagnostic assistance (\cite{thirunavukarasu2023large}), co-pilots for programming (\cite{barke2023grounded}), and autonomous agents for research (\cite{boiko2023autonomous}). This section underscores the unique aspects of developing a causal co-pilot with respect to its {\it grounding}, {\it finetuning}, {\it alignment}, {\it human factors}, and {\it evaluation}.

\vspace{-.1in} 
\subsection{Grounding and Structured Decoding: Regulatory and Causal Frameworks}
\label{sec-grounding}
Despite the potential benefits of leveraging LLMs to enhance the design of medical studies, relying on an off-the-shelf LLM as a ``zero-shot'' co-pilots is unlikely to offer reliable assistance to researchers. The examples in Figure \ref{Fig2} required meticulous and iterative prompting of GPT-4 to produce sensible responses. Often, the pretrained LLM generates generic responses lacking sufficient contextualization for the specific study. Consequently, a primary design challenge in developing causal co-pilots is in adapting an LLM to consistently provide relevant responses that genuinely assist researchers in designing the medical study at hand. This involves anchoring its responses in real-world knowledge and ensuring it maintains relevance to the context. Sources of domain-knowledge for grounding the co-pilot include: (i) causal frameworks developed by previous research for the design of medical studies, and (ii) regulatory guidance on the design, data and analysis considerations that should be addressed in such studies. 

{\bf Causal Frameworks.} 
The Causal Roadmap is an explicit, iterative step-by-step guide for investigators to follow in the prespecification of study design and~statistical analysis plans (e.g., \cite{dang2023causal}).~It~addresses~a~wide range of questions and study types within a single framework. Following the Causal Roadmap leads to two outcomes: (i) detailing key elements of a study design likely to yield high-quality RWE; or (ii) reaching an evidence-based conclusion that producing the required level of RWE is presently unfeasible, while also identifying what data is necessary for future credible RWE generation.~We~refer~the reader to \cite{dang2023causal} for a thorough explanation. 

Another popular causal framework is the Target Trial Emulation approach (e.g., proposed in \citet{hernan2016using}). This framework tries to emulate the conditions under which a clinical trial would have been conducted to answer the causal question researchers are trying to address with RWD. Similar to (and compatible with) the Roadmap \cite{dang2023start}, Target Trial Emulation provides structured processes for critiquing medical studies based on RWD to help avoid common methodological pitfalls. This framework was used to critique the OS examples discussed in Section \ref{Sec2} (\cite{dickerman2019avoidable}). 

{\bf Regulatory Guidance.} Regulatory bodies, such as the FDA, Health Canada, and European Medicines Agency, have established frameworks outlining the criteria and standards for RWD in healthcare and RWE evaluation in regulatory decision-making. These guidelines encompass aspects like data quality, reliability, relevance, and the ethical implications of data use. There are also  specific guidelines dedicated to the data or study types, such as externally controlled trials, non-interventional studies, data standards, and EHR and claims data. By incorporating these regulatory principles into LLMs, researchers can ensure that the generated insights adhere to the highest standards of scientific rigor and ethical compliance. This alignment not only bolsters the credibility and acceptability of the causal co-pilot but also offers the potential to~facilitate~smoother regulatory review processes. Moreover, grounding LLMs in these guidelines enhances their ability to provide contextually relevant, regulation-compliant advice, thereby optimizing the design, execution, and analysis of medical studies using RWD. This integration forms a critical bridge between cutting-edge AI technology and the stringent regulatory landscape governing medical research, paving the way for more effective, efficient, and ethically sound healthcare innovations. 

Future research can investigate how causal and regulatory frameworks can be used to ground causal co-pilots and enforce a structured response to improve the reliability of LLM output. Causal frameworks can be used to ground the co-pilot by using the structured processes defined by these frameworks to systematically refine the user defined prompts and decoding constraints in order to improve the responses of the underlying LLM. These frameworks can also be used to design system prompts that determine the overall behavior of the causal co-pilot. For instance, one can inject a structured description of the different steps of the causal roadmap or the trial emulation framework. LLM decoding programs or queries can also imnprove reliability of the causal co-pilot by enforcing these steps and the careful, structured reasoning they suggest (\cite{beurer2023prompting}). FDA guidance documents and reports can also be used as a knowledge base for retrieval augmentation methods that interface with the causal co-pilot to resurface design and analysis considerations relevant to the study at hand (\cite{shuster2021retrieval}).

\vspace{-.1in} 
\subsection{Finetuning: Instruction-Tuning Data for the Causal Co-Pilot}
\label{train-data} 
While grounding a general-purpose LLM using external sources of knowledge on study design can improve its utility in assisting researchers in designing medical studies, finetuning the LLM itself using high-quality training data for the problem of medical study design might be a more effective way to steer the LLM to function as causal co-pilots (\cite{nori2023can, hernandez2023we}) and reduce ``hallucinations'' in research tasks related to medical study design. Ideally, we would like the causal co-pilot to learn from high-quality ``instruction-tuning'' data, containing demonstrations of medical studies, researcher instructions, and ideal responses from the co-pilot. Similar instruction-tuning datasets are used to train LLMs, such as InstructGPT and FLAN-T5 (\cite{chung2022scaling}) as well as multimodal models such as LLaVA (\cite{liu2023visual}).

However, generating high-quality instruction-tuning data for the causal co-pilot is a challenging task because such data are not naturally occurring or collected in practice. A common approach to alleviate this problem is to use a ``self-instruct'' method: an approach which uses machine-generated, instruction-following data from a state-of-the-art instruction-tuned teacher LLMs (\cite{wang2022self, peng2023instruction}). For instance, the LLaVA model in \cite{liu2023visual} was training using instruction-tuning data generated using a language-only GPT-4 by synthetically processing image captions to generate question-answer pairs associated with each image.

Developing methods for generating instruction-tuning data for the causal co-pilot is an interesting direction for future research. This is a much more~challenging~task~than the self-instruct approaches, where the instruction following data was tailored to common-sense question answering and visual reasoning tasks that easily can be automatically generated or annotated by non-experts. On the contrary, synthetic instruction following datasets that demonstrate the steps involved in the study design process must be grounded to a medical context and are non-trivial to generate. A potential approach for generating these datasets is to use the causal frameworks and regulatory guidance discussed in Section \ref{sec-grounding} to develop structured simulators of synthetic user-LLM interactions that encode knowledge about study design. 

\vspace{-.1in} 
\subsection{Alignment: Learning from Peer Review} 
One driver behind the recent success of LLMs is reinforcement learning from human feedback (RLHF) (\cite{ouyang2022training}). In this finetuning approach, the LLM generations are ``aligned'' with human preferences by gathering feedback on various responses to the same prompt. This human~feedback~is then used to create a reward model predicting preferences on new training data, which in turn refines the LLM through further finetuning. While gathering human feedback for general question-answering tasks is feasible through human annotators, obtaining expert feedback on medical study designs is a more challenging and less scalable process. Future research directions could explore alternative sources for constructing reward models that bypass the bottleneck of expert annotations. One approach is to view reward modeling as a scientific ``peer review'' process, where models are trained using peer reviews of submitted papers or regulatory evaluations of FDA applications, thereby aligning causal co-pilots with scientific and regulatory standards.

\vspace{-.1in}
\subsection{Human Factors: Interactions with\\ Causal Co-pilots}
A key research challenge in building a reliable causal co-pilot is creating a productive interaction between~the~human scientists and the co-pilot (\cite{amershi2019guidelines}). Even simple design decisions, such~as~whether~the co-pilot or the human scientists takes the initiative in executing analysis, whether critiques are presented as mere suggestions or definitive warnings, and details of assistance can strongly influence whether the human considers, trusts and validates, or over-relies on LLM outputs (\cite{gu2023data}).  

\vspace{-.1in}
\subsection{Evaluation: Metrics and Benchmarks} 
Another challenge in developing causal co-pilots is evaluating their success. While it is possible to test the medical knowledge that LLMs encode using benchmarks like USMLE tests (\cite{singhal2022large}), there is currently no benchmark for medical study design. Creating a benchmark with truly held-out test cases is challenging because testing data from published medical studies might already be part of the LLM's training data. For instance, it is unclear whether the model responses in Figure \ref{Fig2} represent the its true capabilities or result from memorization of published studies. Future research could explore defining new metrics to assess LLM-guided study design in terms of alignment with regulatory standards and generating synthetic study design tasks for testing in truly~unseen~settings.

\vspace{-.1in}
\subsection{Related Work}
\label{related-work}
{\bf Research agents in other domains.} An example of adapting LLMs as research agents is the {\it co-scientist} framework in \cite{boiko2023autonomous}. The co-scientist is a system powered by GPT-4 that autonomously tackles complex experiments by interacting with various APIs and tools.~While~it streamlines workflows in fields like organic chemistry, the co-scientist model differs from the causal co-pilot. In natural sciences, experiments are confined to well-defined domains and are free from socio-technical biases. Medical study designs, on the other hand, involve unique contexts, domain-specific knowledge, and highly complex and imperfect data with potential biases, large numbers of latent factors and unmeasured determinants and errors. As a result, we suggest that an effective causal co-pilot for RWE generating in health will not serve to automate or streamline operation, but to collaborate with researchers through an interactive thought process throughout the study design.

{\bf LLMs for causal inference and data analysis.} Another line of work that is closely related to the causal co-pilot examines the capabilities of LLMs in engaging in causal reasoning. For instance, \citet{kiciman2023causal} explores the capabilities of LLMs in various causal reasoning tasks and finds particular promise in LLMs as a source of domain knowledge to augment human domain experts in determining appropriate causal assumptions. \citet{vashishtha2023causal} builds on this work, using LLMs for~identifying~causal estimands for causal effect inference. \cite{lengerich2023llms} further finds that LLMs are capable of interpreting correlational relationships within datasets and linking them to potential mechanisms and identifying surprises. {\color{black}{While the focus on causal reasoning capabilities of LLMs represents a distinct research direction from our proposed agenda, advancements in this area will inevitably impact our agenda.}} A causal co-pilots may build on such capabilities to identify causal assumptions critical to observational studies, as well as potential biases, and systematic measurement errors that may threaten validity of studies.

\vspace{-.1in}
\subsection{Potential Limitations and Harms}
\label{potential-harms}
While an LLM-powered research pipeline holds promising prospects, there exists a potential downside to relying too heavily on systems that are not fully explainable or transparent in our research practices. Erroneous but convincing outputs from LLMs may lead to inaccuracies in defining the causal estimand or identifying relevant~confounders.~This in turn could result in causal identification failures and significant gaps in the estimated causal quantities of interest. These errors could lead to negative downstream impacts, including inflated type 1 and type 2 errors, misleading inference, and an inability to derive actionable insights. Additionally, there is a risk of failing to detect real post-market safety signals or falsely detecting them. Such negative outcomes could result in the underuse of beneficial drugs,~leading~to suboptimal regulatory decisions, clinical support, and individual patient decision-making.

To assess the potential harms stemming from systems like the envisioned co-pilot, we advocate for user studies and randomized trials. These trials are vital for evaluating the utility and limitations of these systems in realistic, complex, and human-centric research environments. Such trials are already being used to evaluate the utility of comparable systems such as Github copilot \cite{dakhel2023github}.

\vspace{-.1in}
\section{Conclusion}
\label{Sec5}
The integration of LLMs as causal co-pilots in the workflow of medical research offers a promising solution to alleviate the burden on investigators and enhance the quality and transparency of medical studies using RWD. The causal co-pilot could bring a contextual depth to study design. Anchoring it in causal inference frameworks and regulatory guidelines can help to ensure human investigator teams use methodologically sound research practices and transparently report them. Advancing this agenda involves tackling unique challenges in pretraining, finetuning, and evaluating the co-pilots. It exemplifies how LLMs can be applied in a highly specialized and impactful domain, showcasing the potential of AI to advance medical research and support regulatory decision-making. 

\section*{Acknowledgments}
This work was supported by Microsoft Accelerating Foundation Models Research (AFMR) Program.

\bibliography{refs}
\bibliographystyle{icml2022}

\newpage
\onecolumn
\section*{Supplementary Appendix}
\subsection*{Experimental Details}
All experiments where conducted using the API for GPT-4-Vision 0125-Preview provided within the Azure Open AI studio. We generated our demonstrative examples using GPT-4 with a temperature parameter of 0.7, a maximum of 800 tokens per response and a ``Top P'' parameter of 0.95. For each experiment, we created a ``system prompt'' that describes how the model should react to each new user prompt. In what follows, we provide samples of system prompts used to generate the model responses in Figures 3C and 3D of the main paper.  

\subsection*{Sample 1}
\vspace{.1in}

\begin{mybox}
{\color{black}{{\bf System Prompt:} 

You are an AI assistant tool that helps statisticians design and analyze the results of observational studies. Whenever the researcher shows you the results of the study (i.e., in the form of survival or cumulative incidence curves) do the following:

1- Describe the scientific figure or table provided so that the researcher ensures your understanding of the results is correct. 

2- Check for evidence of residual confounding and inform the researcher if you have found any. 

Please provide decisive answers and keep your responses concise. Do not provide generic answers that are not related to the specific question provided by the researcher.

}}
\end{mybox} 

\begin{mybox}
{\color{black}{{\bf User Prompt:} 

Can you help me asses the validity of the results from this observational study?

\includegraphics[width=2.5in]{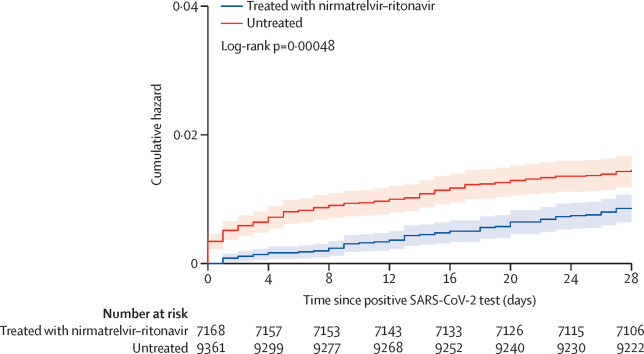}

}}
\end{mybox} 

\begin{mybox}
{\color{black}{{\bf GPT-4 Response:} 

The figure is a cumulative hazard plot comparing two groups over time since a positive SARS-CoV-2 test: one treated with nirmatrelvir-ritonavir and one untreated. The x-axis represents time in days, and the y-axis represents the cumulative hazard. The treated group is represented by the black line, while the untreated group is represented by the red line. The shaded areas around the lines likely represent confidence intervals. The number of individuals at risk in each group at various time points is listed below the graph.  
   
The log-rank p-value is 0.00048, indicating a statistically significant difference in the cumulative hazard between the two groups over the 28-day period.  
   
\hl{Regarding residual confounding, the figure alone does not provide enough information to assess this.} To evaluate residual confounding, we would need additional information on how well the study controlled for potential confounding variables, such as demographics, comorbidities, and other treatments received. Without this information, we cannot conclusively determine whether residual confounding has been adequately addressed.

}}
\end{mybox} 

As we can see, the model is unable to determine whether there are signs of residual confounding in the result provided by the user. In the next sample, we incorporate typical signs of confounding in survival estimates within the system prompt and test whether the model can use this information to assess the existence of residual confounding on a new user-provided input. 

\subsection*{Sample 2}
\vspace{.1in}

\begin{mybox}
{\color{black}{{\bf System Prompt:} 

You are an AI assistant tool that helps statisticians design and analyze the results of observational studies. Whenever the researcher shows you the results of the study (i.e., in the form of survival or cumulative incidence curves) do the following:

1- Describe the scientific figure or table provided so that the researcher ensures your understanding of the results is correct. 

2- Check for evidence of residual confounding and inform the researcher if you have found any. \hl{One way to check for residual confounding is by looking at the survival results for treatment and control groups and assessing whether the timing at which survival curves separate is earlier than expected given the mechanism of action of the intervention being studied. For instance, if the curves separate at time 0 and the treatment is expected to work after 3 days, then this can be an indication of residual confounding.}

Please provide decisive answers and keep your responses concise. Do not provide generic answers that are not related to the specific question provided by the researcher.

}}
\end{mybox} 

\begin{mybox}
{\color{black}{{\bf User Prompt:} 

Can you help me asses the validity of the results from this observational study?

\includegraphics[width=2.5in]{figures/Lancet_study.jpeg}

}}
\end{mybox} 

\begin{mybox}
{\color{black}{{\bf GPT-4 Response:} 

The figure provided is a cumulative hazard plot comparing two groups over time since a positive SARS-CoV-2 test: one that was treated with nirmatrelvir-ritonavir and one that was untreated. The plot shows that the cumulative hazard for the untreated group is consistently higher than for the treated group across the 28-day period, indicating a lower event rate (such as hospitalization, severe disease, or death) in the treated group. The separation between the curves appears to start at day 0 and continues to widen over time. The log-rank p-value is 0.00048, suggesting that the difference in cumulative hazard between the two groups is statistically significant.  
   
\hl{In terms of residual confounding, there is a potential indication of it because the separation between the curves starts immediately at time 0. Since Nirmatrelvir-Ritonavir (a treatment for COVID-19) would be expected to take 3-5 days to exert its effect, an immediate separation could suggest that there are other differences between the treated and untreated groups that are influencing the outcome, not just the treatment itself.} This could include factors such as the severity of the disease at baseline, other treatments being used, or differences in patient characteristics (like age, comorbidities, etc.) that are not accounted for in the analysis.

}}
\end{mybox} 

Here, the model is able to use the general indication of residual confounding described in the system message to reason about the results provided by the user given the known mechanisms of Nirmatrelvir-Ritonavir. In the following sample, we provide another exemplary system message through which the model operates on the same user prompt. 

\newpage
\subsection*{Sample 3}
\vspace{.1in}

\begin{mybox}
{\color{black}{{\bf System Prompt:} 

You are an AI assistant tool that helps statisticians design and analyze the results of observational studies. Whenever the researcher shows you the results of the study (i.e., in the form of survival or cumulative incidence curves) check for evidence of residual confounding and inform the researcher if you have found any. One way to check for residual confounding is by looking at the survival results for treatment and control groups and assessing whether the timing at which survival curves separate is earlier than expected given the mechanism of action of the intervention being studied. For instance, if the curves separate at time 0 and the treatment is expected to work after 3 days, then this can be an indication of residual confounding.

Please provide a decisive answer in one sentence and keep your responses concise. Do not provide generic answers that are not related to the specific question provided by the researcher.

}}
\end{mybox} 

\begin{mybox}
{\color{black}{{\bf User Prompt:} 

Can you help me asses the validity of the results from this observational study?

\includegraphics[width=2.5in]{figures/Lancet_study.jpeg}

}}
\end{mybox} 

\begin{mybox}
{\color{black}{{\bf GPT-4 Response:} 

\hl{There is evidence of residual confounding as the survival curves separate at time 0, which is earlier than expected if the treatment (nirmatrelvir-ritonavir) needs time to take effect.} 

}}
\end{mybox} 

In the following sample, we show the system prompt and model response for the study results shown in Figure 3D.

\subsection*{Sample 4}
\vspace{.1in}
\begin{mybox}
{\color{black}{{\bf System Prompt:} 

You are an AI assistant tool that helps statisticians design and analyze the results of observational studies. Whenever the researcher shows you the results of the study (i.e., in the form of survival or cumulative incidence curves) check for evidence of residual confounding and inform the researcher if you have found any. One way to check for residual confounding is by looking at the survival results for treatment and control groups and assessing whether the timing at which survival curves separate is earlier than expected given the mechanism of action of the intervention being studied. 

Please provide a decisive answer in one sentence and keep your responses concise. Do not provide generic answers that are not related to the specific question provided by the researcher.

}}
\end{mybox} 

\newpage
\begin{mybox}
{\color{black}{{\bf User Prompt:} 

Can you help me asses the validity of the results from this observational study?

\includegraphics[width=4.5in]{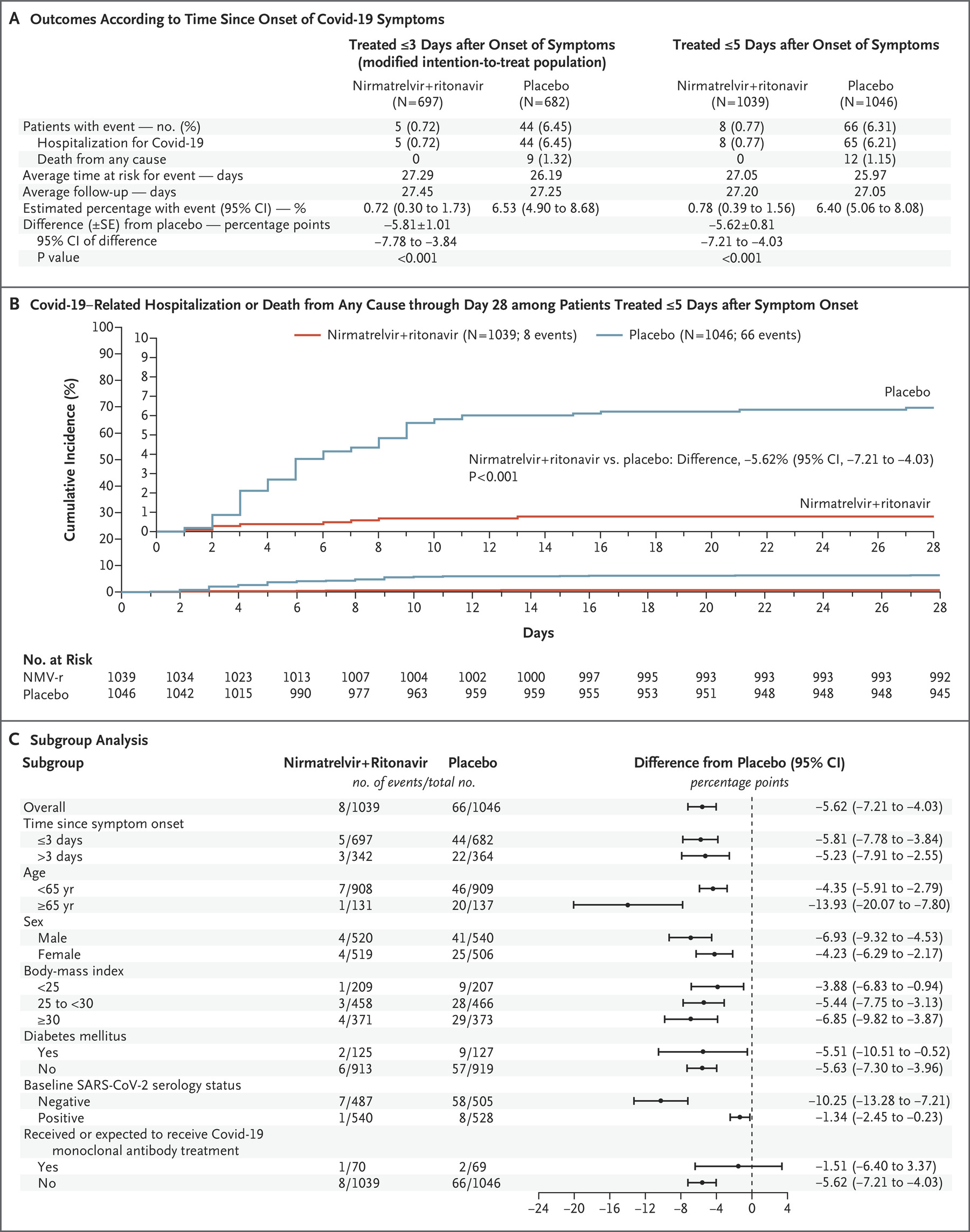}

}}
\end{mybox} 

\begin{mybox}
{\color{black}{{\bf GPT-4 Response:} 

The provided image is from a randomized controlled trial, not an observational study, as indicated by the presence of a placebo group and the assignment of treatments to participants. Therefore, the results would not typically be subject to the same concerns about residual confounding that are relevant to observational studies.

}}
\end{mybox} 

In this example, the model was able to recognize that the data comes from an RCT rather than an observational study as the user mistakenly mentioned. It did so by recognizing the existence of a placebo group in the analysis. The model also realizes that residual confounding is not a concern since the data comes from a randomized experiment.

\end{document}